\documentclass{article}

\usepackage[round]{natbib}
\usepackage[preprint]{neurips_data_2022}

\usepackage{bbm}
\usepackage[utf8]{inputenc} 
\usepackage[T1]{fontenc}    
\usepackage{hyperref}       
\usepackage{url}            
\usepackage{amsmath}
\usepackage{booktabs}       
\usepackage{amsfonts}       
\usepackage{nicefrac}       
\usepackage{microtype}      
\usepackage{xcolor}         
\usepackage{xspace}
\usepackage{todonotes}
\usepackage[etex=true,export]{adjustbox}
\usepackage{authblk}    
\usepackage[capitalise]{cleveref}
\usepackage{makecell}
\usepackage{subcaption}
\usepackage{multirow}
\usepackage{epsfig}
\usepackage{epstopdf}
\usepackage{bbding}
\newtheorem{theorem}{Theorem}
\usepackage{makecell}

\newcommand{\chat}{ChatGPT\xspace}
\newcommand{\advglue}{AdvGLUE\xspace}
\newcommand{\flip}{Flipkart\xspace}
\newcommand{\ddx}{DDXPlus\xspace}
\newcommand{\error}[1]{\textcolor{red}{#1}}

\title{On the Robustness of ChatGPT: An Adversarial and Out-of-distribution Perspective}

\author{%
  \textbf{Jindong Wang}$^{1}$\thanks{Contact: jindong.wang@microsoft.com.}, \textbf{Xixu Hu}$^{1,2\ddagger}$\thanks{Equal contribution.}, \textbf{Wenxin Hou}$^{3\dagger}$, \textbf{Hao Chen}$^{4}$, \textbf{Runkai Zheng}$^{1,5}$\thanks{Work done during internship at Microsoft Research Asia.}, \textbf{Yidong Wang}$^{6}$, \textbf{Linyi Yang}$^{7}$, \textbf{Wei Ye}$^{6}$, \textbf{Haojun Huang}$^{3}$, \textbf{Xiubo Geng}$^{3}$, \textbf{Binxing Jiao}$^{3}$, \textbf{Yue Zhang}$^{7}$, \textbf{Xing Xie}$^{1}$ 
}
\affil{\small{$^{1}$Microsoft Research, $^{2}$City University of Hong Kong, $^{3}$Microsoft STCA, $^{4}$Carnegie Mellon University,\\$^{5}$Chinese University of Hong Kong (Shenzhen), $^{6}$Peking University, $^{7}$Westlake University}

\url{https://github.com/microsoft/robustlearn}
}

\begin{document}

\maketitle

\begin{abstract}

\chat is a recent chatbot service released by OpenAI and is receiving increasing attention over the past few months.
While evaluations of various aspects of \chat have been done, its robustness, i.e., the performance to unexpected inputs, is still unclear to the public.
Robustness is of particular concern in responsible AI, especially for safety-critical applications.
In this paper, we conduct a thorough evaluation of the robustness of \chat from the adversarial and out-of-distribution (OOD) perspective.
To do so, we employ the \advglue and ANLI benchmarks to assess adversarial robustness and the Flipkart review and \ddx medical diagnosis datasets for OOD evaluation.
We select several popular foundation models as baselines.
Results show that \chat shows consistent advantages on most adversarial and OOD classification and translation tasks.
However, the absolute performance is far from perfection, which suggests that adversarial and OOD robustness remains a significant threat to foundation models.
Moreover, \chat shows astounding performance in understanding dialogue-related texts and we find that it tends to provide informal suggestions for medical tasks instead of definitive answers.
Finally, we present in-depth discussions of possible research directions.

\end{abstract}

\section{Introduction}
\label{sec-intro}

Large language models (LLMs), or foundation models~\citep{bommasani2021opportunities}, have achieved significant performance on various natural language process (NLP) tasks.
Given their superior in-context learning capability~\citep{min2022rethinking}, prompting foundation models has emerged as a widely adopted paradigm of NLP research and applications.
\chat is a recent chatbot service released by OpenAI~\citep{chatgpt}, which is a variant of the Generative Pre-trained Transformers (GPT) family.
Thanks to its friendly interface and great performance, \chat has attracted over $100$ million users in two months.

It is of imminent importance to evaluate the potential risks behind \chat given its increasing worldwide popularity in diverse applications.
While previous efforts have evaluated various aspects of \chat in law~\citep{choi2023chatgpt}, ethics~\citep{shen2023chatgpt}, education~\citep{khalil2023will}, and reasoning~\citep{bang2023multitask}, we focus on its \emph{robustness}~\citep{bengio2021deep}, which, to our best knowledge, has not been thoroughly evaluated yet.
Robustness refers to the ability to withstand disturbances or external factors that may cause it to malfunction or provide inaccurate results.
It is important to practical applications especially the safety-critical scenarios.
For instance, if we apply \chat or other foundation models to fake news detection, a malicious user might add noise or certain perturbations to the content to bypass the detection system.
Without robustness, the reliability of the system collapses.

Robustness threats exist in a wide range of scenarios: out-of-distribution (OOD) samples~\citep{wang2022generalizing}, adversarial inputs~\citep{goodfellow2014explaining}, long-tailed samples~\citep{zhang2021deep}, noisy inputs~\citep{natarajan2013learning}, and many others.
In this paper, we pay special attention to two popular types of robustness: the adversarial and OOD robustness, both of which are caused through input perturbation.
Specifically, adversarial robustness studies the model's stability to adversarial and imperceptible perturbations, e.g., adding trained noise to an image or changing some keywords of a text.
On the other hand, OOD robustness measures the performance of a model on unseen data from different distributions of the training data, e.g., classifying sketches using a model trained for art painting or analyzing a hotel review using a model trained for appliance review.
More background of these robustness is elaborated in \cref{sec-back-robust}.

\textbf{Zero-shot robustness evaluation.}
While robustness research often requires training and optimization (e.g., fine-tuning, linear probing, domain adaptation and generalization, \cref{sec-back-robust}), in this work, we focus on \emph{zero-shot} robustness evaluation.
Given a foundation model, we perform inference directly on the test dataset for evaluation.
We argue that it becomes more expensive and unaffordable to train, or even load existing (and future, larger) foundation models.
For instance, the largest Flan-T5 model has $11$ billion parameters~\citep{chung2022scaling}, which is already beyond the capability of most researchers and practitioners.
Thus, their zero-shot performance becomes important to downstream tasks.
On the other hand, foundation models are typically trained on huge volumes of datasets with huge amount of parameters, which seems to challenge conventional machine learning theories (\cref{append-theory}):
\begin{center}
    \emph{Are large foundation models all we need for robustness?}
\end{center}

\begin{figure}[t!]
    \centering
    \includegraphics[width=\textwidth]{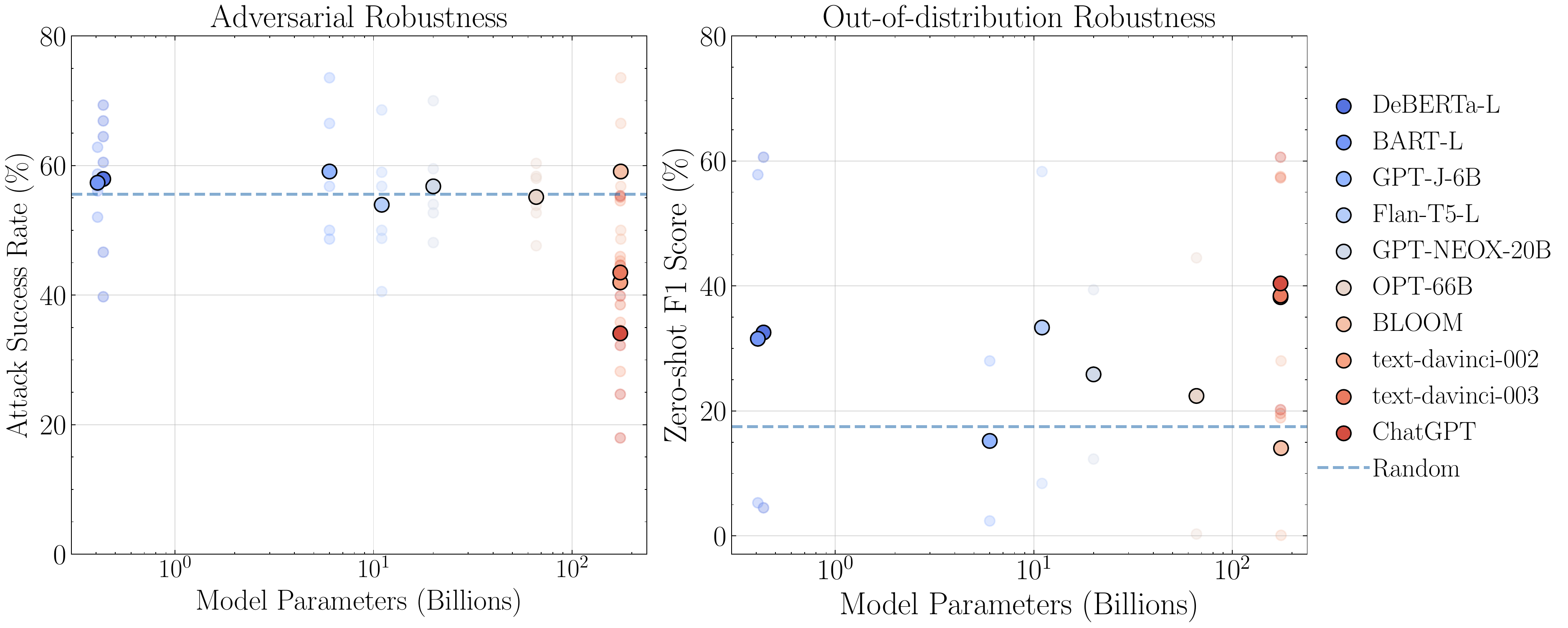}
    \caption{Robustness evaluation of different foundation models: performance vs. parameter size. Results show that \chat shows consistent advantage on adversarial and OOD classification tasks. However, its absolute performance is far from perfection, indicating much room for improvement.}
    \label{fig-summary}
\end{figure}

In this work, we conduct a thorough evaluation of \chat on its adversarial and OOD robustness for natural language understanding tasks.
It is challenging to select appropriate datasets for evaluating 
 \chat since it is known to be trained on huge text datasets as of 2021.
Eventually, we leverage several recent datasets for our evaluation: \advglue~\citep{wang2021adversarial} and ANLI~\citep{nie2019adversarial} for adversarial robustness and two new datasets for OOD robustness: \flip review~\citep{flipkart_2023} and \ddx medical diagnosis datasets~\citep{tchango2022ddxplus}.
Furthermore, we randomly selected $30$ samples from \advglue to form an adversarial translation dataset to evaluate the translation performance.
These datasets represent various levels of robustness, thus provide a fair evaluation.
The detailed information of these datasets are introduced in \cref{sec-dataset}.
We then select several popular foundation models from Huggingface model hub and OpenAI service\footnote{Huggingface: \url{https://huggingface.co/models}. OpenAI service: \url{https://openai.com/api}.} to compare with \chat.
In summary, we have $9$ tasks and overall $2,089$ test examples.\footnote{Although the sample size may seem small, it must be noted that due to the current unavailability of an API service for \chat, conducting experiments on a larger scale is challenging.} 

\textbf{Our findings.}
We perform zero-shot inference on all tasks using these models and \cref{fig-summary} summarizes our main results.
The major findings of the study include:
\begin{enumerate}
    \item What \chat does well: 
    \begin{itemize}
        \item \chat shows consistent improvements on most adversarial and OOD classification tasks.
        \item \chat is good at translation tasks. Even in the presence of adversarial inputs, it can consistently generate readable and reasonable responses.
        \item \chat is better at understanding dialogue-related texts than other foundation models. This could be attributed to its enhanced ability as a chatbot service, leading to good performance on \ddx dataset.
    \end{itemize}
    
    \item What \chat does not do well:
    \begin{itemize}
        \item The absolute performance of \chat on adversarial and OOD classification tasks is still far from perfection even if it outperforms most of the counterparts.
        \item The translation performance of \chat is worse than its instruction-tuned sibling model text-davinci-003.
        \item \chat does not provide definitive answers for medical-related questions, but instead offers informed suggestions and analysis. Thus, it can serve as a friendly assistant.
    \end{itemize}


    \item Other general findings about foundation models:
    \begin{itemize}
        \item Task-specific fine-tuning helps language models perform better on related tasks, e.g., NLI-fine-tuned RoBERTa-L has similar performance to Flan-T5-L.
        \item Instruction tuning benefits large language models, e.g., Flan-T5-L achieves comparable performance to text-davinci-002 and text-davinci-002 with significantly less parameters.
    \end{itemize}
        
\end{enumerate}

Beyond evaluations, we share more reflections in the discussion and limitation sections, providing experience and suggestions to future research.
Finally, we open-source our code and results at \url{https://github.com/microsoft/robustlearn} to facilitate future explorations.

\section{Background}
\label{sec-back}

\subsection{Foundation Models, \chat, and Existing Evaluation}

Foundation models have become a popular research and application paradigm for natural language process tasks.
Since foundation models are trained on large volumes of data, they show significant performance improvement on different downstream tasks such as sentiment analysis, question answering, automatic diagnosis, logical reasoning, and sequence tagging.
\chat is a generative foundation model that belongs to the GPT-3.5 series in OpenAI's GPT family, coming after GPT~\citep{radford2018improving}, GPT-2~\citep{radford2019language}, GPT-3~\citep{brown2020language}, and InstructGPT~\citep{ouyang2022training}.
In contrast to its predecessors, \chat makes it easy for every one to use just through a browser with enhanced multi-turn dialogue capabilities.
Although the technical details of \chat is still not released, it is known to be trained using reinforcement learning from human feedback (RLHF)~\citep{christiano2017deep} with instruction tuning.
Other than natural language processing, there are also emerging efforts in building foundation models for computer vision~\citep{dehghani2023scaling}, music generation~\citep{agostinelli2023musiclm}, biology~\citep{luo2022biogpt,lee2020biobert}, and speech recognition~\citep{radford2022robust}.

Previous efforts evaluate \chat in different aspects~\citep{van2023chatgpt}.
\citet{bang2023multitask} proposes a multi-task, multi-modal, and multilingual evaluation of \chat on different tasks.
They showed that \chat performs reasonably well on most tasks, while it does not bring great performance on low-resource tasks.
Similar empirical evaluations are also made by \citet{gozalo2023chatgpt,azaria2022chatgpt}.
Specifically, \citet{qin2023chatgpt} also did several evaluations and they found that \chat does not do well on fine-grained downstream tasks such as sequence tagging.
In addition to research from artificial intelligence, researchers from other areas also showed interest in \chat.
\citet{hacker2023regulating,shen2023chatgpt} expressed concerns that \chat and other large models should be regulated since they are double-edged swords.
The evaluations on ethics are done in \citep{zhuo2023exploring}.
There are reflections and discussions from law~\citep{choi2023chatgpt}, education~\citep{khalil2023will,m2022exploring,susnjak2022chatgpt,guo2023close}, human-computer interaction~\citep{tabone2023using}, medicine~\citep{jeblick2022chatgpt}, and writing~\citep{biswas2023chatgpt}.
To the best of our knowledge, a thorough robustness evaluation is currently under-explored.

\subsection{Robustness}
\label{sec-back-robust}

In the following, we present the formulation of robustness with the classification task (other tasks can be formulated similarly).
We are given a $K$-class classification dataset $\mathcal{D}=\{\mathbf{x}_i, y_i\}_{i=1}^n$, where $\mathbf{x} \in \mathbb{R}^d$ and $y \in [K]$ are its $d$-dimensional input and output, respectively.
We use $\ell[\cdot, \cdot]$ to denote the loss function.

\paragraph{Adversarial robustness}
An adversarial input~\citep{goodfellow2014explaining} $\mathbf{x}^\prime$ is generated by adding a $\epsilon$-bounded, imperceptible perturbation $\delta$ to the original input $\mathbf{x}$.
The optimal classifier can be learned by optimizing the following objective~\citep{madry2017towards}:
\begin{equation*}
    \min_{f \in \mathcal{H}} \mathbb{E}_{(\mathbf{x}, y) \in \mathcal{D}} \max_{|\delta| \le \epsilon} \ell [f(\mathbf{x} + \delta), y].
\end{equation*}

\paragraph{Out-of-distribution robustness}
On the other hand, OOD robustness (generalization)~\citep{wang2022generalizing,shen2021towards} aims to learn an optimal classifier on an unseen distribution by training on existing data.
One popular formulation for OOD robustness is to minimize the average risk on all distributions $e$, which is sampled over the set of all possible distributions (could be large than $\mathcal{D}$):
\begin{equation*}
    \min_{f \in \mathcal{H}} \mathbb{E}_{e \sim \mathcal{Q}} \mathbb{E}_{(\mathbf{x}, y) \in \mathcal{D}^e} \ell[f(\mathbf{x}), y].
\end{equation*}

\citet{yang2022glue} presented GLUE-X, a benchmark based on GLUE and then conducted a thorough evaluation of the OOD robustness of language models by training on in-distribution (ID) sets and then testing on OOD sets.
Ours, however, performs zero-shot evaluation.
The OOD robustness of \chat cannot be evaluated by GLUE and GLUE-X benchmarks since it may include the entire GLUE datasets in its training data.

\section{Datasets and Tasks}
\label{sec-dataset}

\subsection{Adversarial Datasets}

We adopt \advglue~\citep{wang2021adversarial} and adversarial natural language inference (ANLI)~\citep{nie2019adversarial} benchmarks for evaluating adversarial robustness.
\advglue is a modified version of the existing GLUE benchmark~\citep{wang2019glue} by adding different kinds of adversarial noise to the text: word-level perturbation (typo), sentence-level perturbation (distraction), and human-crafted perturbations.
We adopt $5$ tasks from \advglue: SST-2, QQP, MNLI, QNLI, and RTE.
Since the test set of \advglue is not public, we adopt its development set instead for evaluation.
Although \advglue is a classification benchmark, we additionally construct an adversarial machine translation (En $\to$ Zh) dataset,  termed \advglue-T, by randomly selecting $30$ samples from \advglue. 

ANLI is a large-scale dataset designed to assess the generalization and robustness of natural language inference (NLI) models, which was created by Facebook AI Research. It comprises 16,000 premise-hypothesis pairs that are classified into three categories: entailment, contradiction, and neutral. The dataset is divided into three parts (R1, R2, and R3) based on the number of iterations used during its creation, with R3 being the most difficult and diverse.
Therefore, we select the test set of R3 for evaluating the adversarial robustness of our models. Detailed information of \advglue and ANLI can be found in \cref{append-data-adv}.

\begin{table}[t!]
\caption{Statistics of test sets in this paper}
\label{tb-dataset}
\centering
\begin{tabular}{ccccc}
\toprule
Area & Dataset & Task & \#Sample & \#Class \\ \midrule
\multirow{7}{*}{\begin{tabular}[c]{@{}c@{}}Adversarial   \\ robustness\end{tabular}} & SST-2 & sentiment classification & 148 & 2 \\ 
 & QQP & quora question pairs & 78 & 3 \\ 
 & MNLI & multi-genre natural language inference & 121 & 3 \\ 
 & QNLI & question-answering NLI & 148 & 2 \\ 
 & RTE & textual entailment recognition & 81 & 2 \\ 
 & ANLI & text classification & 1200 & 3 \\
 & \advglue-T & machine translation (En $\to$ Zh) & 30 & - \\ 
 \midrule
\multirow{2}{*}{\begin{tabular}[c]{@{}c@{}}OOD \\ robustness\end{tabular}} & Flipkart & sentiment classification & 331 & 2 \\ 
 & DDXPlus & medical diagnosis classification & 100 & 50 \\ \bottomrule
\end{tabular}
\end{table}

\subsection{Out-of-distribution Datasets}

We adopt two new datasets\footnote{Considering \chat is reported to be trained on a substantial corpus of internet language data as of 2021, identifying an out-of-distribution dataset poses a difficulty. Furthermore, we concern that previous natural language processing datasets predating 2022 may have been assimilated by \chat, so we only utilize datasets that are recently released.} for OOD robustness evaluation: \flip~\citep{flipkart_2023} and \ddx~\citep{tchango2022ddxplus}.
\flip is a product review dataset and \ddx is a new medical diagnosis dataset, both of which are released in 2022.
These two datasets can be used to construct classification tasks.
We randomly sample a subset of each dataset to form the test sets.
Detailed introduction and construction of each test set can be found in \cref{append-data-ood}.
\cref{tb-dataset} shows the statistics of each dataset.

\emph{Remark:}
Finding an OOD dataset for large models like \chat is difficult due to the unavailability of its training data.
Consider these datasets as `out-of-example' datasets since they did not show up in \chat's training data.
Additionally, distribution shift may happen at different dimensions: not only across domains, but also across time.
Thus, even if \chat and other LLMs may already use similar datasets like medical diagnosis and product review, our selected datasets are still useful for OOD evaluation due to temporal distribution shift.
Finally, we must admit the limitation of these datasets and look forward to brand new ones for more thorough evaluation.

\section{Experiment}

\subsection{Zero-shot Classification}

\subsubsection{Setup}

We compare the performance of \chat on AdvGLUE classification benchmark with the following existing popular foundation models:
DeBERTa-L~\citep{he2020deberta}, BART-L~\citep{lewis2020bart}, GPT-J-6B~\citep{gpt-j},
Flan-T5~\citep{raffel2020exploring,chung2022scaling}, 
GPT-NEOX-20B~\citep{neox20b},
OPT-66B~\citep{zhang2022opt}, 
BLOOM~\citep{scao2022bloom}, and GPT-3 (text-davinci-002 and text-davinci-003)~\footnote{Note that the classification task may be unfavorable to the generative models since we did not limit their output space as discriminative models like DeBERTa-L do.}.
The latter two are from OpenAI API service and the rest are on Huggingface model hub.
The notation `-L' means `-large', as we only evaluate the large version of these models.
The detailed information of these models are introduced in \cref{append-model}.

For adversarial classification tasks on \advglue and ANLI, we adopt attack success rate (ASR) as the metric for robustness.
The metric details are listed in \cref{append-metrics-asr}.
For OOD classification tasks, F1-score (F1) is adopted as the metric.
As mentioned before, we only perform zero-shot evaluation.
Thus, we simply run all models on a local computer with plain GPUs, which could be the case in most downstream applications.\footnote{Even the local computer is not that ``plain'' since it requires at least $1$ A$100$ GPU with $80$ GB of memory.}
Note that we use the NLI-fine-tuned version of DeBERTa-L and BART-L on natural language inference tasks to perform zero-shot classification since they are not originally designed for text classification.
For other models, we adopt the prompt-based paradigm to get answers for classification by inputting prompts.
All prompts used in this paper are presented in \cref{append-promp}.
Note that we manually processed some outputs since the outputs of some generative LLMs are not easy to control.

\begin{table}[t!]
\centering
\caption{Zero-shot classification results on adversarial (ASR$\downarrow$) and OOD (F1$\uparrow$) datasets. The best and second-best results are highlighted in \textbf{bold} and \underline{underline}.}
\label{tb-results}
\resizebox{\textwidth}{!}{
\begin{tabular}{l|cccccc|cc}
\toprule
\multirow{2}{*}{Model \& \#Param.} & \multicolumn{6}{c|}{Adversarial robustness (ASR$\downarrow$)} & \multicolumn{2}{c}{OOD robustness (F1$\uparrow$)} \\ 
 & \multicolumn{1}{c}{SST-2} & \multicolumn{1}{c}{QQP} & \multicolumn{1}{c}{MNLI} & \multicolumn{1}{c}{QNLI} & \multicolumn{1}{c}{RTE} & \multicolumn{1}{c|}{ANLI} & \multicolumn{1}{c}{Flipkart} & \multicolumn{1}{c}{\ddx}  \\ \midrule
Random & \multicolumn{1}{c}{50.0} & \multicolumn{1}{c}{50.0} & \multicolumn{1}{c}{66.7} & \multicolumn{1}{c}{50.0} & \multicolumn{1}{c}{50.0} & \multicolumn{1}{c}{66.7} & \multicolumn{1}{c}{20.0} & \multicolumn{1}{c}{4.0}  \\ \midrule
DeBERTa-L (435 M) & \multicolumn{1}{c}{66.9} & \multicolumn{1}{c}{39.7} & \multicolumn{1}{c}{64.5} & \multicolumn{1}{c}{46.6} & \multicolumn{1}{c}{60.5} & \multicolumn{1}{c|}{69.3} & \multicolumn{1}{c}{\textbf{60.6}} & \multicolumn{1}{c}{4.5} \\ 
BART-L (407 M) & \multicolumn{1}{c}{56.1} & \multicolumn{1}{c}{62.8} & \multicolumn{1}{c}{58.7} & \multicolumn{1}{c}{52.0} & \multicolumn{1}{c}{56.8} & \multicolumn{1}{c|}{\underline{57.7}}  & \multicolumn{1}{c}{57.8} & \multicolumn{1}{c}{5.3}  \\ \midrule
GPT-J-6B (6 B) & \multicolumn{1}{c}{48.7} & \multicolumn{1}{c}{59.0} & \multicolumn{1}{c}{73.6} & \multicolumn{1}{c}{50.0} & \multicolumn{1}{c}{56.8} & \multicolumn{1}{c|}{66.5}  & \multicolumn{1}{c}{28.0} & \multicolumn{1}{c}{2.4}  \\ 
Flan-T5-L (11 B) & \multicolumn{1}{c}{\underline{40.5}} & \multicolumn{1}{c}{59.0} & \multicolumn{1}{c}{48.8} & \multicolumn{1}{c}{50.0} & \multicolumn{1}{c}{56.8} & \multicolumn{1}{c|}{68.6}  & \multicolumn{1}{c}{58.3} & \multicolumn{1}{c}{8.4}  \\ 
GPT-NEOX-20B (20 B) & \multicolumn{1}{c}{52.7} & \multicolumn{1}{c}{56.4} & \multicolumn{1}{c}{59.5} & \multicolumn{1}{c}{54.0} & \multicolumn{1}{c}{48.1} & \multicolumn{1}{c|}{70.0}  & \multicolumn{1}{c}{39.4} & \multicolumn{1}{c}{12.3}  \\ 
OPT-66B (66 B) & \multicolumn{1}{c}{47.6} & \multicolumn{1}{c}{53.9} & \multicolumn{1}{c}{60.3} & \multicolumn{1}{c}{52.7} & \multicolumn{1}{c}{58.0} & \multicolumn{1}{c|}{\underline{58.3}} & \multicolumn{1}{c}{44.5} & \multicolumn{1}{c}{0.3} \\ 
BLOOM (176 B) & \multicolumn{1}{c}{48.7} & \multicolumn{1}{c}{59.0} & \multicolumn{1}{c}{73.6} & \multicolumn{1}{c}{50.0} & \multicolumn{1}{c}{56.8} & \multicolumn{1}{c|}{66.5}  & \multicolumn{1}{c}{28.0} & \multicolumn{1}{c}{0.1} \\ 
text-davinci-002 (175 B) & \multicolumn{1}{c}{46.0} & \multicolumn{1}{c}{\underline{28.2}} & \multicolumn{1}{c}{54.6} & \multicolumn{1}{c}{45.3} & \multicolumn{1}{c}{35.8} & \multicolumn{1}{c|}{68.8}  & \multicolumn{1}{c}{57.5} & \multicolumn{1}{c}{18.9}  \\ 
text-davinci-003 (175 B) & \multicolumn{1}{c}{44.6} & \multicolumn{1}{c}{55.1} & \multicolumn{1}{c}{\underline{44.6}} & \multicolumn{1}{c}{\underline{38.5}} & \multicolumn{1}{c}{\underline{34.6}} & \multicolumn{1}{c|}{62.9}  & \multicolumn{1}{c}{57.3} & \multicolumn{1}{c}{\underline{19.6}}  \\ 
ChatGPT (175 B) & \multicolumn{1}{c}{\textbf{39.9}} & \multicolumn{1}{c}{\textbf{18.0}} & \multicolumn{1}{c}{\textbf{32.2}} & \multicolumn{1}{c}{\textbf{34.5}} & \multicolumn{1}{c}{\textbf{24.7}} & \multicolumn{1}{c|}{\textbf{55.3}}  & \multicolumn{1}{c}{\textbf{60.6}} & \multicolumn{1}{c}{\textbf{20.2}} \\ \bottomrule
\end{tabular}
}
\end{table}

\subsubsection{Results}

The classification results of adversarial and OOD robustness are shown in \cref{tb-results}.

First, \textbf{\chat shows consistent improvements on adversarial datasets.}
It outperforms all counterparts on all adversarial classification tasks.
However, we see that there is still room for improvement since the absolute performance is far from perfection.
For instance, the ASRs on SST-2 and ANLI are $40\%$ and $55.3\%$, respectively, indicating much room for improvement. 
This could be due to the reason that they are trained on clean corpus and some adversarial texts are washed out from the training data.
Beyond \chat, it is also surprising to find that most methods only achieve slightly better than random guessing, while some even do not beat random guessing.
This indicates that the zero-shot adversarial robustness of most foundation models is not promising.
Such adversarial vulnerability poses a major threat to various applications of foundation models, which we will further discuss in \cref{sec-discuss-adv} and \cref{append-theory-ml}.
In addition to foundation models, we are surprised to find that some small models also achieve great performance on adversarial tasks while it has much less parameters than the strong models (e.g, DeBERTa-L on QQP and QNLI tasks).
This indicate that fine-tuning on relevant tasks can still improve the performance.
Furthermore, Flan-T5 also achieves comparable performance to most larger models.
Since Flan-T5 is also trained via instruction tuning, this implies the efficacy of such training approach in prompting-based NLP tasks.

Second, \textbf{all models after GPT-2 (text-davinci-002, text-davinci-003, and \chat) perform well on OOD datasets.}
This observation is in consistency with recent finding in OOD research that the in-distribution (ID) and OOD performances are positively correlated~\citep{miller2021accuracy}.
However, \chat and its sibling models perform much better on \ddx, indicating its ability to recognize new or diverse domain data.
Additionally, some large models performs better, e.g., Flan-T5-L outperforms some larger models such as OPT-66B and BLOOM.
This can be explained as overfitting on certain large models or they have an \emph{inverse} ID-OOD relation~\citep{teney2022id} on our test sets.
It should also be noted that the absolute performance of \chat and davinci series are still far from perfection.
More discussions on OOD are presented in \cref{sec-discuss-ood} and \cref{append-data-ood} shows some informal analysis from the perspective of OOD theory.

Third, on the \ddx dataset, \textbf{\chat is better at understanding diaglogue-related texts compared with other LLMs.}
The \ddx benchmark presents a formidable challenge for many models. The majority of models perform at a level akin to random chance, with the exception of the davinci series and \chat, which exhibit exceptional performance.
One plausible explanation for the superior performance of these three models may be their substantial increase in the number of model parameters.
This substantial increase in parameter count may enable the model to learn more complex representations and subsequently result in an improvement of performance.
Another possible reason for the success of \chat is its ability to understand the conversational context of \ddx, which consists of doctor-posed diagnostic questions and patient responses.
\chat has demonstrated superior performance in understanding conversational context compared to previous models, which likely contributes to its improved performance on this dataset.

Finally, it is worth noting that due to the critical nature of the healthcare field, \textbf{\chat does not provide definitive answers in medical-related questions but instead offers informed suggestions and analysis, followed by a recommendation for further offline testing and consultation to ensure accurate diagnosis.}
When the provided information is insufficient to make a judgment, \chat will acknowledge this and offer an explanation, demonstrating its responsible approach to medical-related inquiries.
This highlights the benefits of using \chat for medical-related inquiries compared to search engines, as it can provide comprehensive analysis and suggestions without requiring the users to have medical expertise, while also being responsible and cautious in its responses.
This suggests a promising future for the integration of \chat in computer-aided diagnosis systems.

\subsection{Zero-shot Machine Translation}

\subsubsection{Setup}
We further evaluate the adversarial robustness of ChatGPT on an English-to-Chinese (En $\to$ Zh) machine translation task.
The test set (\advglue-T) is sub-sampled from the adversarial English text in \advglue and we manually translate them into Chinese as ground truth.
We evaluate the zero-shot translation performance of \chat against text-davinci-002 and text-davinci-003.
We further adopt two fine-tuned machine translation models from the Huggingface model hub: OPUS-MT-EN-ZH~\citep{tiedemannThottingalEAMT2020} and Trans-OPUS-MT-EN-ZH\footnote{Note that there are only few En $\to$ Zh machine translation models released on Huggingface model hub and we pick the top two with the most downloads.}
More details of the models used are included in \cref{append-model}.
We report BLEU, GLEU, and METEOR in experiments to conduct a fair comparison among several models.\footnote{We use NLTK (\url{https://www.nltk.org/}) to calculate these metrics.}

\subsubsection{Results}

The results of zero-shot machine translation are shown in \cref{tab:translation}. 
Note that all three models from the GPT family outperforms the fine-tuned models. 
Interestingly, text-davinci-003 generalizes the best on all metrics. 
The performance of ChatGPT is better to text-davinci-002 on BLUE and GLUE, but slightly worse on METOR. 
While differing in metrics, we find \textbf{the translated texts of ChatGPT (and text-davinci-002 and text-davinci-003) is very readable and reasonable to humans, even given adversarial inputs.}
This indicates the adversarial robustness capability on machine translation of ChatGPT might originate from GPT-3.

\begin{table}[htbp]
\centering
\caption{Zero-shot machine translation results on adversarial text sampled from AdvGLUE.}
\label{tab:translation}
\resizebox{0.5\columnwidth}{!}{%
\begin{tabular}{@{}c|ccc@{}}
\toprule
 Model & BLEU$\uparrow$ & GLEU$\uparrow$ & METOR$\uparrow$ \\ \midrule
OPUS-MT-EN-ZH & 18.11 & 26.78 & 46.38 \\
Trans-OPUS-MT-EN-ZH & 15.23 & 24.89 & 45.02 \\
text-davinci-002 & 24.97 & 36.30 & \underline{59.28} \\ 
text-davinci-003 & \textbf{30.60} & \textbf{40.01} & \textbf{61.88} \\ 
\chat & \underline{26.27} & \underline{37.29} & 58.95 \\ 
 \bottomrule
\end{tabular}%
}
\end{table}

\subsection{Case Study}

\cref{tb-adv-example} shows some results of \chat across word-level (typo) and sentence-level (distraction) adversarial inputs.
It is evident that both adversaries pose a considerable challenge to \chat, through their ability to mislead the model's judgement. It should be noted that these adversaries are prevalent in everyday interactions, and the existence of numerous forms of textual adversarial attacks highlights the necessity of defensive strategies for \chat.
\cref{tb-ood-example} presents some cases of \chat on OOD inputs.
Unlike adversarial inputs, it is not easy to analyze why \chat performs bad for OOD datasets since the notion of ``distribution'' is hard to quantify.


\begin{table}[t!]
\caption{Case study on adversarial examples. Adversarial manipulations are marked \error{red}.}
\label{tb-adv-example}
\resizebox{\textwidth}{!}{
\begin{tabular}{c| p{7cm} |c|c|c}
\toprule
\multicolumn{1}{c|}{Type} & \multicolumn{1}{c|}{Input} & Truth & davinci003 & \chat \\ \midrule
 & i think \error{you 're} here for raunchy college humor . & Positive & Negative & Negative \\ \cmidrule{2-5} 
 & Mr. Tsai is a very \error{oriignal} artist in his medium , and what time is it there? & Positive & Positive & Positive \\ \cmidrule{2-5} 
 & \makecell[l]{Q1: Can you TRANSLATE these to English \\language?  \\ Q2: \error{Cn} you translate \error{ths} from Bengali to English \\ \error{lagnuage}?} & \makecell[c]{Not \\ equivalent} & \makecell[c]{Not \\ equivalent} & Equivalent \\ \cmidrule{2-5} 
\multirow{-4}{*}{\makecell[c]{word-level\\ (typo)}} & \makecell[l]{Q1: What are the best things in \error{Hog} Kong? \\ Q2: What is the best thing in Hong Kong?} & Equivalent & \makecell[c]{Not \\ equivalent} & \makecell[c]{Not \\ equivalent} \\ \midrule
 & \makecell[l]{Question: What is the minimum \error{required} if you \\ want to teach in Canada? \\ Sentence: \error{@KMcYo0} In most provinces a second \\ Bachelor's Degree such as a Bachelor of Education \\ is required to become a qualified teacher.} & \makecell[c]{Not \\ entailment} & Entailment & Entailment \\ \cmidrule{2-5} 
 & \makecell[l]{Question: \error{@uN66rN} What kind of water body is \\ rumored to be obscuring Genghis Khan's burial \\ site? \\ Sentence: Folklore says that a river was diverted \\ over his grave to make it impossible to find (the \\ same manner of burial as the Sumerian King \\ Gilgamesh of Uruk and Atilla the Hun).} & Entailment & \makecell[c]{Not \\ entailment} & \makecell[c]{Not \\ entailment} \\ \cmidrule{2-5} 
 & \error{https://t.co/1GPp0U} the iditarod lasts for days - this just felt like it did . & Negative & Positive & \makecell[c]{Negative} \\ \cmidrule{2-5} 
\multirow{-4}{*}{\makecell[c]{sentence-level\\ (distraction)}} & holden caulfield did it better .  \error{https://t.co/g4vJKP} & Negative & Positive & Negative \\ \bottomrule
\end{tabular}
}
\end{table}

\section{Discussion}


\subsection{Adversarial Attack Remains a Major Threat}
\label{sec-discuss-adv}

As discussed in experiments, dealing with adversarial inputs still remains challenging to large foundation models.
With the proliferation of foundation model service such as \chat, such adversarial vulnerability remains a major threat to various downstream scenarios, especially those safety-critical applications.
On the other hand, since adversarial inputs are subjectively generated by humans, but not exist in nature, we argue that foundation models might never cover all distributions of possible adversarial inputs during their training~\citep{ilyas2019adversarial}.
Other than error correction, a possible solution for model owners is to first inject adversarial inputs to their training data, which could improve its robustness to existing adversarial noise.
Then, as a long-standing goal to improve the model robustness, the pre-trained model can be continuously trained on human-generated or algorithm-generated adversarial inputs.

As for those who cannot train large models and only use them in downstream tasks, such threat still exists due to the defect inheritance of pre-trained models.
In this case, how to achieve perfect fine-tuning or adaptation performance on downstream tasks while certainly reducing the defect inheritance remains a major challenge.
Luckily, some pioneering work~\citep{zhang2022remos, chin2021renofeation} might provide solutions.
This represents a novel and emerging direction for future research.
However, as foundation models grow larger that go beyond the capabilities of most researchers, reducing the defects through fine-tuning could be impossible.
An open question rises for both model owners and downstream users on how to defend the adversarial attack.

In addition to adversaries in training data, prompts can also be attacked~\citep{maus2023adversarial}, which requires further knowledge and algorithms to deal with.
This is currently a challenging problem due to the sensitivity of prompting to LLMs.

\subsection{Can OOD Generalization be Solved by Large Foundation Models?}
\label{sec-discuss-ood}

Larger models like \chat and text-davinci-003 have the potential to achieve superior performance on OOD datasets with better prompt engineering, inspiring us to think of the problem: is OOD generalization solved by these giant models?
The huge training data and parameters are a double-edged sword: overfitting vs. generalization.
It is also intuitive to think that OOD data is unseen during training, so adding it into training set is enough, which is what these large models did.
Is the ``unreasonable effectiveness of data''~\citep{sun2017revisiting} real?
However, as the model sizes are becoming larger, it still remains unknown when and why LLMs will overfit.

Another possible reason is the training data of \chat and text-davinci-003 actually encompass similar distributions to our test sets even if they are collected after 2021.
\flip is for product review and \ddx is for medical diagnosis, which in fact are common domains widely existing on the Internet.
Thus, they could be not OOD to these models, that could lead to overfitting.
New datasets from long-tailed domains are in need for more fair evaluations.

Finally, our analysis does not show that ID-OOD performances are always positively correlated~\citep{miller2021accuracy}, but can sometimes inversely correlated~\citep{teney2022id}.
Regularization and other techniques should be developed to improve the OOD performance of language models.

\subsection{Beyond NLP Foundation Models}

Adversarial and OOD robustness do not only exist in natural language, but also in other domains.
In fact, most research comes from machine learning and computer vision communities.
Researchers in computer vision area could possibly think: can we solve OOD and adversarial robustness in image data by training a vision foundation model?
For instance, the recent ViT-22B~\citep{dehghani2023scaling} scales vision Transformer~\citep{dosovitskiy2020image} to 22 billion parameters by training it on the 4 billion JFT dataset~\citep{zhai2022scaling} (a larger version of the previous JFT-300M dataset~\citep{sun2017revisiting}), which becomes the largest vision foundation model to date.
ViT-22B shows superior performance on different image classification tasks.
However, it does not show ``emergent abilities''~\citep{wei2022emergent} with the increment of parameters as other LLMs.
Not only LLMs, the robustness in other areas also remains to be solved.

Back to theory, algorithms, and optimization areas, which foundational research areas in artificial intelligence.
Will the large foundation models disrupt these areas?
First, we should acknowledge that the success of foundation models should also attribute to these areas, e.g., most LLMs adopt the Transformer~\citep{vaswani2017attention} and other advanced learning and training research.
Second, the success of foundation models shed light on these areas: can we solve the problems like adversarial and OOD by developing new theories, algorithms, and optimization methods?
Such research could offer valuable contribution to foundation models, e.g., improve the data and training efficiency and efficacy.
Finally, researchers in these areas should not be dis-encouraged since the advance of scientific research should be diverse and not restricted to those done with rich computing resources.


\section{Limitation}

This paper offers a preliminary empirical study on the robustness of large foundation models, which has the following limitations.

First, we only perform zero-shot classification using \chat and other models.
Results of these models could change if we perform fine-tuning or adaptation given enough computing resources.
But as we stated in introduction, it is expensive and un-affordable to perform further operation on today's latest foundation models, we believe zero-shot evaluation is reasonable.

Second, it seems controversial to evaluate large foundation models on small datasets in this work.
However, since the training data of \chat and some large models remains unclear, it is difficult to find larger datasets.
Especially, \chat is trained on huge datasets on the Internet as of 2021, making it more difficult to find appropriate datasets for thorough evaluation.
We do believe more datasets can be used for such evaluation.

Third, we did most evaluations on text classification and only minor evaluations on machine translation.
It is well-known that \chat and other foundation models can do more tasks such as generation.
Again, because of lack of appropriate datasets, evaluating generation performance is also difficult.
We also admit that introducing more proper prompts could improve its performance.

Fourth, it is worth noting that \chat is mainly designed to be a chatbot service rather than a tool for text classification.
Our evaluations are mainly for classification, which have nothing to do with the robustness of \chat for online chatting experience.
We do hope every end-user can find \chat helpful in their lives.

Finally, we could further provide detailed synopsis by conducting experiments on data before 2021 as comparisons and analyzing more OOD cases to see why \chat succeeds or fails.
Other experiments include detailed ablation study using different language models and investigation of induced outputs by LLMs through prompts.
These can be done in future work.
Another claim is that \chat is not perfect for adversarial tasks.
But we also need to develop certain metrics to show `how good' is the performance.

\section{Conclusion}

This paper presented a preliminary evaluation of the robustness of \chat from the adversarial and out-of-distribution perspective.
While we acknowledge the advance of large foundation models on adversarial and out-of-distribution robustness, our experiments show that there is still room for improvement to \chat and other large models on these tasks.
Afterwards, we presented in-depth analysis and discussion beyond NLP area, and then highlight some potential research directions regarding foundation models.
We hope our evaluation, analysis, and discussions could provide experience to future research.

\section*{Acknowledgement}

This paper received attentions from many experts since its first version was released on ArXiv.
Authors would like to thank all who gave constructive feedback to this work.

\section*{Disclaimer}

\paragraph{Potential Ethics and Societal Concerns raised by \chat Robustness}
The increasing popularity of \chat and other chatbot services certainly face some new concerns from both ethics and society.
The purpose of this paper is to show that \chat can be attacked by adversarial and OOD examples using existing public dataset, but not to attack it intentionally.
We hope that this will not be leverage by end-users.
Finally, we also hope the community can realize the importance of robustness research and develop new technologies to make our systems more secure, robust, and responsible.

\paragraph{\chat usage}
Some authors in this paper are from mainland China where \chat is currently unavailable.
In order to conduct this research without disobeying local laws and OpenAI service terms, Hao Chen, who is one of our coauthors and lives in U.S., did all experiments related to \chat and OpenAI.
All experiments on \chat are based on its Feb 13 version.
Further updates of \chat may lead to change of the results in this paper.

\paragraph{The contribution of each author}
Jindong led the project, designed experiments, wrote the code framework, and wrote the paper.
Xixu and Wenxin shared equal contributions.
Xixu was in charge of processing, experimenting, and writing about the \ddx and ANLI datasets.
Wenxin designed all prompts to generative models and wrote about this part.
Hao did the machine translation experiments, wrote necessary codes, and was in charge of code organization and reproducibility.
Runkai helped polish the paper and organized case study.
Other authors actively participated in this project from day one, reviewed the paper carefully, and provided valuable comments to improve this work.

\bibliography{refs}
\bibliographystyle{plainnat}

\newpage

\appendix

\section{Detailed Introduction of Datasets and Tasks}

\subsection{\advglue and ANLI}
\label{append-data-adv}

\advglue~\citep{wang2021adversarial} is an evaluation benchmark for natural language processing models, with a specific focus on adversarial robustness. It includes five natural language understanding tasks from the GLUE benchmark: Sentiment Analysis (SST-2), Duplicate Question Detection (QQP), and Natural Language Inference (NLI, including MNLI, RTE, QNLI).
It includes different types of attacks including word-level transformations, sentence-level manipulations, and human-written adversarial examples.

\textbf{SST-2} The Stanford Sentiment Treebank~\citep{socher2013recursive} is composed of sentences originating from movie reviews, along with corresponding human-annotated sentiments. The goal is to predict the sentiment (positive or negative) when given a review sentence.

\textbf{QQP} Quora Question Pairs (QQP) dataset consists of pairs of questions gathered from Quora, which is a platform for community question-answering. The aim is to predict if two questions are semantically equivalent.

\textbf{MNLI} Multi-Genre Natural Language Inference Corpus~\citep{williams2018broad} is a dataset of sentence pairs for textual entailment. The task is to predict whether the premise sentence entails, contradicts, or is neutral to the hypothesis sentence.

\textbf{QNLI} The Question-answering NLI (QNLI) dataset consists of question-sentence pairs extrated and modified from the Stanford Question Answering Dataset~\citep{rajpurkar2016squad}. The task is to predict if the context sentence has the answer to a given question.

\textbf{RTE} The Recognizing Textual Entailment (RTE) dataset contains examples constructed using news and Wikipedia text from annual textual entailment challenges. The goal is to predict the relationship between a pair of sentences, which can be categorized into two classes: entailment and not entailment. Note that neutral and contradiction are considered as not entailment.

\textbf{\advglue-T} We create an adversarial machine translation dataset (En $\to$ Zh) called \advglue-T by randomly extracting 30 samples from \advglue. 

\textbf{Adversarial NLI (ANLI)}~\citep{nie2020adversarial} is a benchmark for natural language understanding collected by using human-and-model-in-the-loop training method. This benchmark is designed to challenge the current models in natural language inference. Human annotators acted as adversaries by trying to fool the model into mis-classifying with the found vulnerabilities, while these sentences are still understandable to other humans.

\subsection{\flip and \ddx}
\label{append-data-ood}

\flip~\citep{flipkart_2023} includes information on 104 different types of products from {\ttfamily flipkart.com}, such as electronics, clothing, home decor, and more. It contains 205,053 data and their corresponding sentiment labels (positive, negative, or neutral).
In our study, we select all its instances with review text length between $150$ and $160$ to ease the experiments.
This leads to $331$ samples in total.

\ddx~\citep{tchango2022ddxplus} is a dataset designed for automatic medical diagnosis, which consists of synthetic data of around $1.3$ million patients, providing a differential diagnosis and the true pathology, symptoms, and antecedents for each patient. We randomly sampled 100 records from the test set. As the original records were in French, we translated them into English using the evidences and conditions dictionaries provided in the dataset. The resulting data was then formatted into a context of age, gender, initial evidence, and inquiry dialogue, enabling the model to select the most probable disease from all considered pathology using the information provided in the conversation.

\section{Evaluation Metrics} \label{append-metrics-asr}
\paragraph{Attack Success Rate (ASR)} 
Following \citep{wang2021adversarial}, the metric of ASR is adopted for evaluating the effectiveness of the system against adversarial inputs. Specifically, given a dataset $\mathcal{D} = \{(x_i, y_i)\}_{i=1}^N$ consisting of $N$ samples $x_i$ and corresponding ground truth labels $y_i$, the success rate of an adversarial attack method $\mathcal{A}$, which generates adversarial examples $A(x)$ given an input $x$ to attack a surrogate model $f$, is computed as: 

\begin{equation}
\operatorname{ASR} = \sum\limits_{(x,y)\in \mathcal{D}} \frac{\mathbbm{1}[f(\mathcal{A}(x)) \neq y]}{\mathbbm{1}[f(x) = y]}
\end{equation}

Basically, the robustness of a model is inversely proportional to the attack success rate. 

\section{An Informal Analysis from the Theory Perspective}
\label{append-theory}

This section presents a brief overview of existing machine learning and robustness theory, assisting potential analysis of large foundation models.

\subsection{Machine Learning Theory}
\label{append-theory-ml}

The foundational learning theory in machine learning is called the probably approximately correct (PAC) theory~\citep{valiant1984theory}.
While our focus is to facilitate the analysis of foundation models, we only discuss the theory related to generalization error, which is the basic one.

In binary classification, we define the true labeling function $f: \mathcal{X} \rightarrow [0,1]$ for domain $\mathcal{D}$. For any classifier $h:\mathcal{X} \rightarrow [0,1]$, the classification error is defined as:
\begin{equation}
    \epsilon(h,f) = \mathbb{E}_{x \sim \mathcal{D}}[h(x) \neq f(x)] = \mathbb{E}_{x \sim \mathcal{D}}[|h(x) - f(x)|].
\end{equation}

\begin{theorem}[Generalization error]
\label{theory-generalization}
Let $\mathcal{H}$ be a finite hypothesis set, $m$ the number of training samples, and $0 < \delta < 1$, then for any $h \in \mathcal{H}$,
\begin{equation}
    P \left( |\mathbb{E}(h) - \hat{\mathbb{E}}(h)| \le \sqrt{\frac{\ln |\mathcal{H}| + \ln (2/\delta)}{2m}} \right) \ge 1 - \delta,
\end{equation}    
where $\mathbb{E}(h)$ and $\hat{\mathbb{E}}(h)$ are the ideal and empirical (learned) risk on $h$, respectively.
\end{theorem}

\cref{theory-generalization} indicates that the generalization error is determined by the number of training samples $m$ and the size of the hypothesis space $\mathcal{H}$.
The superior performance of large foundation models are typically trained on huge datasets ($m$ is large).
However, the hypothesis set $\mathcal{H}$ is finite.
Therefore, the increment of $m$ and $|\mathcal{H}|$ could lead to a lower generalization error according to \cref{theory-generalization}.
This seems to explain why large foundation models such as \chat and text-davinci-003 achieve superior performance in zero-shot classification on some tasks.
Note taht the theoretical analysis on foundation models is still underexplored, hence, this analysis could be wrong and we still look forward to theoretical advances in this area.

However, as large foundation models become more complex, it could possibly induce a high VC-dimension~\citep{valiant1984theory}.
At the same time, their training data sizes are certainly larger than existing machine learning research.
It remains unknown why such models do not overfit on existing datasets.

\subsection{Out-of-distribution Robustness Theory}
\label{append-theory-ood}

OOD assumes training on a source dataset $\mathcal{D}_s$ and test on another unseen dataset $\mathcal{D}_t$.
The key challenge is that the distributions between $\mathcal{D}_s$ and $\mathcal{D}_t$ are not the same.
Although it is impossible to evaluate the risk on an unseen dataset since we cannot even access it, we can borrow the classic domain adaptation theory to analyze the risk on the target domain by assuming its availability.

\begin{theorem}[Target error bound based on $\mathcal{H}$-divergence~\citep{ben2010theory}]
\label{theo:h}
Let $\mathcal{H}$ be a hypothesis space with VC dimension $d$. Given sample set with size $m$ i.i.d. sampled from the source domain, then, with probability at least $1-\delta$, for any $h \in \mathcal{H}$, we have:
\begin{equation}
    \epsilon_t(h) \leq \hat{\epsilon}_s(h)  + d_{\mathcal{H}}(\hat{\mathcal{D}}_s, \hat{\mathcal{D}}_t) + \lambda^\ast + \sqrt{\frac{4}{m} \left(d\log\frac{2em}{d}+\log\frac{4}{\delta} \right)},
\end{equation}
where $e$ is natural logarithm, $\lambda^\ast = \epsilon_s(h^\ast) + \epsilon_t(h^\ast)$ is the ideal joint risk, and $h^\ast = \mathop{\arg\min}\limits_{h \in \mathcal{H}} \epsilon_s(h) + \epsilon_t(h)$ is the optimal classifier on the source and target domains.
\end{theorem}

Theory \ref{theo:h} indicates that the error bound on the target domain is bounded by four terms: 1) source empirical error, 2) the distribution discrepancy between source and target domains, 3) ideal joint error, and 4) some constant related to sample size and VC dimension.

Conventional OOD generalization and adaptation research~\citep{wang2022generalizing} focus on minimizing the distribution discrepancy between source and target domains ($d_{\mathcal{H}}(\hat{\mathcal{D}}_s, \hat{\mathcal{D}}_t)$) while assuming the source risk ($\hat{\epsilon}_s(h)$) is determined.
Meanwhile, the last term ($\sqrt{\cdot}$) can also be reduced due to the increment of $m$.
Similar to the above generalization analysis, we can also interpret the success of large foundation models as they simply achieving low generalization error on the source data, thus also minimizes the risk on the target domain.
But it is also important to note that this analysis is not rigorous.
Finally, VC-dimension has no correlation with the distribution of datasets, which also cannot explain the strong OOD performance of these foundation models.

\section{Foundation Models used in Experiments}
\label{append-model}

In this section, we provide a brief introduction to the foundation models used in our experiments.




\textbf{BART-L~\citep{lewis2020bart}} BART is based on bidirectional and auto-regressive transformer. It is trained on a combination of auto-regressive and denoising objectives, which makes BART feasible for both generation and understanding tasks. In a nutshell, BART is designed to handle both understanding and generation tasks, making it a more versatile model, while BERT is more focused on understanding.


\textbf{DeBERTa-L~\citep{he2020deberta}} DeBERTa introduces a disentangled attention mechanism and an enhanced decoding scheme for BERT. The disentangled attention mechanism allows DeBERTa to capture the contextual information between different tokens in a sentence more effectively, while the enhanced decoding scheme makes the model generate natural language sentences with higher quality. 


\textbf{GPT-J-6B}~\citep{gpt-j} is a transformer model trained using Mesh Transformer JAX~\citep{mesh-transformer-jax}. It is a series of models with `6B' denoting 6 billion parameters.

\textbf{Flan-T5~\citep{raffel2020exploring,chung2022scaling}} Flan-T5 adopts a text-to-text strategy where input and output are both natural language sentences to execute a variety of tasks like machine translation, summarization, and question answering. This input-output form allows Flan-T5 to accomplish held-out tasks when given an input sentence as prompt.


\textbf{GPT-NEOX-20B~\citep{neox20b}} GPT-NeoX-20B is a language model with 20 billion parameters trained on the Pile. It is the largest public dense autoregressive model. It outperformed GPT-3 and FairSeq models with similar size in five-shot reasoning tasks.

\textbf{OPT~\citep{zhang2022opt}} Open Pre-trained Transformers (OPT) is a suite of pre-trained transformer models that are decoder-only and have parameter sizes ranging from 125 million to 175 billion. While offering comparable performance to GPT-3~\citep{brown2020language}, OPT-175B was developed with just 1/7th of the carbon footprint.

\textbf{BLOOM~\citep{scao2022bloom}} BLOOM extends pre-training from mono-lingual to cross-lingual. BLOOM combines one unsupervised objective and one supervised objective for pre-training. The unsupervised one only uses monolingual data, and the supervised one adopts parallel data. The cross-lingual language models can bring significant improvements for low-resource languages.

\textbf{text-davinci-002 and text-davinci-003} text-davinci-002 and text-davinci-003~\footnote{\url{https://platform.openai.com/docs/models/gpt-3}} are based on GPT-3~\citep{brown2020language}. They accomplish any task that other models can, generally produce output that is of higher quality, longer in length, and more faithful to instructions.

\section{Details on Prompts}
\label{append-promp}

\subsection{Prompts}
We list all prompts used in this study in \cref{tb-prompt}.

\begin{table}[htbp]
\caption{All prompts used in this study.}
\label{tb-prompt}
\resizebox{\textwidth}{!}{
\begin{tabular}{l|p{12cm}}
\toprule
Dataset & Prompt \\ \midrule
SST-2 & {\ttfamily Please classify the following sentence into either positive or negative. Answer me with "positive" or "negative", just one word.} \\ \midrule
QQP & {\ttfamily Are the following two questions equivalent or not? Answer me with "equivalent" or "not\_equivalent".} \\ \midrule
MNLI & {\ttfamily Are the following two sentences entailment, neutral or contradiction? Answer me with "entailment", "neutral" or "contradiction".} \\ \midrule
QNLI & {\ttfamily Are the following question and sentence entailment or not\_entailment? Answer me with "entailment" or "not\_entailment".} \\ \midrule
RTE & {\ttfamily Are the following two sentences entailment or not\_entailment? Answer me with "entailment" or "not\_entailment".} \\ \midrule
AdvGLUE-T & {\ttfamily Translate the following sentence from Engilish to Chinese.} \\ \midrule
ANLI & {\ttfamily Are the following paragraph entailment, neutral or contradiction? Answer me with "entailment", "neutral" or "contradiction". The answer should be a single word. The answer is:} \\ \midrule
\flip & {\ttfamily Is the following sentence positive, neutral, or negative? Answer me with "positive", "neutral", or "negative", just one word.} \\ \midrule
\ddx & {\ttfamily Imagine you are an intern doctor. Based on the previous dialogue, what is the diagnosis? Select one answer among the following lists: {[}'spontaneous pneumothorax', 'cluster headache', 'boerhaave', 'spontaneous rib fracture', 'gerd', 'hiv (initial infection)', 'anemia', 'viral pharyngitis', 'inguinal hernia', 'myasthenia gravis', 'whooping cough', 'anaphylaxis', 'epiglottitis', 'guillain-barré syndrome', 'acute laryngitis', 'croup', 'psvt', 'atrial fibrillation', 'bronchiectasis', 'allergic sinusitis', 'chagas', 'scombroid food poisoning', 'myocarditis', 'larygospasm', 'acute dystonic reactions', 'localized edema', 'sle', 'tuberculosis', 'unstable angina', 'stable angina', 'ebola', 'acute otitis media', 'panic attack', 'bronchospasm / acute asthma exacerbation', 'bronchitis', 'acute copd exacerbation / infection', 'pulmonary embolism', 'urti', 'influenza', 'pneumonia', 'acute rhinosinusitis', 'chronic rhinosinusitis', 'bronchiolitis', 'pulmonary neoplasm', 'possible nstemi / stemi', 'sarcoidosis', 'pancreatic neoplasm', 'acute pulmonary edema', 'pericarditis', 'cannot decide'{]}. The answer should be a single word. The answer is:} \\ \bottomrule
\end{tabular}
}
\end{table}

\subsection{OOD Case Study}
\label{append-oodexample}

We list some of the OOD examples for case study in \cref{tb-ood-example}.

\begin{table}[t!]
\caption{Case study on OOD examples.}
\label{tb-ood-example}
\resizebox{\textwidth}{!}{
\begin{tabular}{l|c|c|c}
\toprule
\multicolumn{1}{c|}{Input} & Truth & davinci003 & \chat \\ \midrule
\makecell[l]{quality of cover is not upto mark but the content in the book is \\ really good from foundation to difficult level questions are of latest \\ pattern great work} & Positive & Positive & Positive \\ \midrule
\makecell[l]{worst product dont buy flipcart should not sell such useless product \\ prepared food only one time it damaged smoke came out and burned \\ it good for nothing} & Positive & Negative & Negative \\ \midrule
\makecell[l]{definitely it will not fit wagon r either front or back it will cover one \\ side fully and the other side partially thickness is not that much \\ average product} & Positive & Negative & Negative \\ \midrule
\makecell[l]{this ink is genuine but the problem with printer is it shows red light \\ after 100pages but i still used the cartridge and at last 357 pages \\ were printed} & Negative & Positive & Neutral \\ \midrule
\makecell[l]{working fine good but received in messy box and there is bent on \\ inverter at corner think mistake of courier facility whatever but \\ working fine no issue} & Negative & Positive & Positive \\ \bottomrule
\end{tabular}
}
\end{table}

\end{document}